\documentclass[letterpaper, 10 pt, conference]{ieeeconf}  

\IEEEoverridecommandlockouts                              

\title{\LARGE \bf
RuN: Residual Policy for Natural Humanoid Locomotion
}

\author{Qingpeng Li$^{1}$, Chengrui Zhu$^{1}$, Yanming Wu$^{1}$, Xin Yuan$^{1}$, Zhen Zhang$^{1}$, Jian Yang$^{2}$ and Yong Liu$^{1\dagger}$
\thanks{$^\dagger$ Corresponding Author}
\thanks{$^{1}$ Zhejiang University}
\thanks{$^{2}$ China Research and Development Academy of Machinery Equipment, Beijing, China}
}

\usepackage{graphicx}
\usepackage[utf8]{inputenc}
\usepackage{amssymb}
\usepackage{threeparttable}
\usepackage{amsmath}
\usepackage{amsfonts}
\usepackage{booktabs}
\usepackage{multirow}
\usepackage{subcaption}
\usepackage{dsfont}
\usepackage[table]{xcolor} 

\begin{document}

\pdfminorversion=7

\maketitle
\thispagestyle{empty}
\pagestyle{empty}


\begin{abstract}
Enabling humanoid robots to achieve natural and dynamic locomotion across a wide range of speeds, including smooth transitions from walking to running, presents a significant challenge. Existing deep reinforcement learning methods typically require the policy to directly track a reference motion, forcing a single policy to simultaneously learn motion imitation, velocity tracking, and stability maintenance. To address this, we introduce RuN, a novel decoupled residual learning framework. RuN decomposes the control task by pairing a pre-trained Conditional Motion Generator, which provides a kinematically natural motion prior, with a reinforcement learning policy that learns a lightweight residual correction to handle dynamical interactions. Experiments in simulation and reality on the Unitree G1 humanoid robot demonstrate that RuN achieves stable, natural gaits and smooth walk–run transitions across a broad velocity range (0–2.5 m/s), outperforming state-of-the-art methods in both training efficiency and final performance.
\end{abstract}
\section{introduction}

Humanoid robots are expected to operate in human-centric environments across a broad spectrum of tasks. Among these capabilities, reliable locomotion is the most fundamental and central prerequisite. However, developing a unified locomotion controller that spans a wide range of speeds, from walking to running, and enables smooth transitions between these gaits remains a pivotal challenge.

Classical control methods, such as those based on Zero-Moment Point (ZMP), can ensure stability but often produce rigid, robotic gaits ill-suited for high-speed dynamic tasks~\cite{zmp}. In parallel, Model Predictive Control (MPC) optimizes short-horizon plans under simplified dynamics and preset contact schedules due to real-time and modeling limits, which can restrict agility and robustness on humanoids~\cite{herdt2010online}. In response, Deep Reinforcement Learning (DRL) has emerged as a powerful paradigm for learning complex motor skills end-to-end~\cite{radosavovic2024real,li2025reinforcement}. However, the effectiveness of DRL depends on reward design, and even small misspecification can steer learning toward easy-to-exploit objectives, yielding efficient yet non-anthropomorphic motions rather than structured, human-like behavior.

\begin{figure}[t]
    \vspace{2mm}
    \centering
    \includegraphics[width=\linewidth]{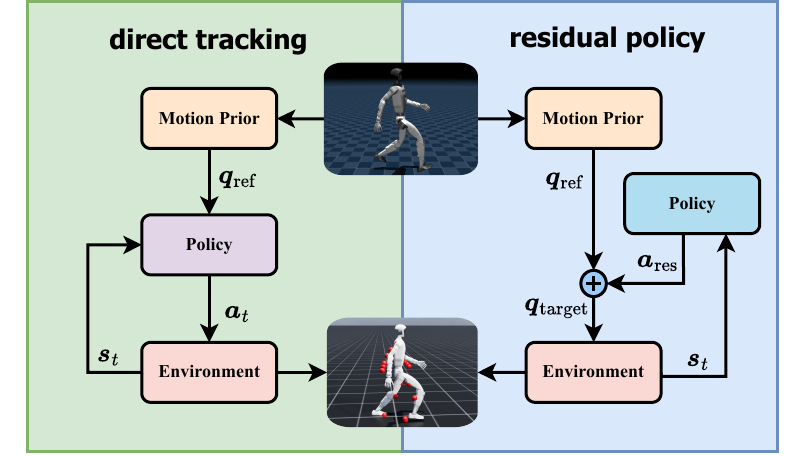}
    \caption{Comparison between direct tracking (left) and our residual policy (right). Rather than directly tracking the motion prior $\boldsymbol{q}_\text{ref}$, our policy outputs a residual action $\boldsymbol{a}_{\text{res}}$. The action applied to the environment is \(\boldsymbol{q}_\text{target}=\boldsymbol{q}_{\text{ref}}+\boldsymbol{a}_{\text{res}}\), simplifying learning and improving motion quality.}
    \label{fig:residual_vs_tracking}
\end{figure}

To promote naturalness in DRL, researchers incorporate data-driven motion priors. Paradigms such as the Generative Motion Prior (GMP)~\cite{gmp} employ generative models to synthesize natural reference motions as priors, feed these priors directly into a reinforcement learning policy, and train the policy with imitation and task rewards to produce natural movements. However, this direct-tracking strategy creates a significant learning-complexity challenge in which a single policy must simultaneously satisfy three often competing objectives—imitating the kinematic style of the motion prior, maintaining dynamic stability, and executing task commands. This tight coupling exposes the policy to a large and complex action space, lengthening training time. Additionally, inherent conflicts among the objectives constrain performance and necessitate trade-offs among motion quality, stability, and task accuracy.

To address the inherent complexities in DRL, we propose RuN, a novel decoupled residual learning framework. The core idea of this framework is to decompose the complex end-to-end control task into two more tractable sub-modules. The first module, a Conditional Motion Generator (CMG), autoregressively generates high-quality kinematic reference trajectories from given velocity commands, which serve as a motion prior. The second module is a residual policy that learns a significantly simplified task, which is providing small and dynamic corrections to the motion prior to compensate for external perturbations. The final control command executed by the robot is the summation of the motion priors generated by the CMG and the residual action output by the policy. This residual formulation drastically narrows the policy's exploration space, allowing it to focus its learning capacity on mastering complex dynamic interactions, thereby enabling our controller to achieve smooth transitions from walking to running.

We conduct extensive experiments on the Unitree G1 humanoid robot. The results demonstrate that our framework trains significantly faster and achieves superior final performance compared to state-of-the-art methods across all key metrics, including motion naturalness and velocity tracking error. Our robot successfully performs natural walking and running across a continuous velocity range of 0–2.5~m/s, with smooth gait transitions in the real world.

Our contributions can be summarized as follows:

\begin{itemize}
\item A novel decoupled residual learning framework, RuN, that simplifies humanoid locomotion by decomposing the action into a motion prior and a learned residual, leading to significant gains in final performance.
\item A novel Conditional Motion Generator that autoregressively produces a range of natural motions from walking to running, serving as a high-quality motion prior for the residual policy.
\item In extensive real-world and simulation trials on the Unitree G1 humanoid robot, our controller, RuN, demonstrates superior performance over state-of-the-art methods regarding motion naturalness and velocity tracking.
\end{itemize}
\section{related work}

\subsection{Learning-based Humanoid Locomotion}

Learning-based approaches, particularly DRL, have significantly advanced humanoid robot locomotion without relying on precise dynamics models. Li et al.~\cite{li2025reinforcement} demonstrated versatile, dynamic, and robust bipedal locomotion control through DRL. Siekmann et al.~\cite{siekmann2021blind} enabled blind bipedal stair traversal without visual perception using sim-to-real reinforcement learning. Zhuang et al.~\cite{zhuang2024humanoid} trained humanoid robots to perform complex parkour movements via DRL. Crowley et al.~\cite{crowley} optimized bipedal locomotion for the 100-meter dash and compared it with human running performance. Haarnoja et al.~\cite{haarnoja2024learning} taught bipedal robots agile soccer skills with DRL, including ball chasing and recovery after falling. These studies highlight the remarkable ability of DRL to address high-dimensional and nonlinear dynamics problems. However, most of these works focus primarily on task accomplishment and dynamic performance, while the generated motions often lack the naturalness of human movements, appearing relatively stiff and mechanical.

\subsection{Data-driven Human Motion Generation}

Data-driven human motion generation has made remarkable progress in recent years, aiming to synthesize natural, diverse, and condition-consistent human pose sequences from signals such as text, audio, or scene context. Guo et al.~\cite{guo2020action2motion} proposed a VAE-based architecture that generates motion sequences conditioned on specified action categories. Petrovich et al.~\cite{petrovich22temos} further extended this approach by incorporating Transformer layers into the VAE, enabling the model to learn a joint distribution of text and motion and thus produce diverse motion sequences from more complex natural language descriptions. Tevet et al.~\cite{tevet2023human} pioneered the use of diffusion models in this domain, directly predicting samples instead of noise at each diffusion step, which significantly improved motion quality and realism. This generative paradigm was soon extended to more complex tasks—for instance, Tseng et al.~\cite{tseng2022edge} applied diffusion models to music-driven dance generation. In contrast, Wang et al.~\cite{wang2022humanise} focused on integrating natural language instructions with 3D scene constraints, producing human motions that are consistent with the environment. It is worth noting that most of these models are primarily designed for offline applications in computer graphics, and using them as motion priors for online robotic control imposes much stricter requirements on computational efficiency and real-time performance.

\subsection{Imitation with Motion Priors}

To enhance the naturalness of generated motions, a large body of work has incorporated human motion data as priors. A classical line of research is imitation learning, exemplified by Peng et al.~\cite{peng2018deepmimic}, which jointly optimizes task objectives and motion imitation. Building on this, Peng et al.~\cite{amp} introduced the Adversarial Motion Prior (AMP), where a discriminator is trained to distinguish between robot motions and human reference motions, thereby providing style rewards for physics-based characters. This framework was later extended to Adversarial Skill Embeddings (ASE)~\cite{peng2022ase}, which learn reusable skill embeddings at scale. Moving towards real robots, Zhang et al.~\cite{zhang2024whole} applied AMP to achieve whole-body humanoid locomotion guided by human reference data. To overcome the limitations of discriminative priors, Zhang et al.~\cite{gmp} proposed Generative Motion Priors, replacing the discriminator with a frozen generative model that outputs interpretable reference trajectories during training. By learning human motion distributions offline with a conditional VAE and providing dense, fine-grained supervision signals to the policy during online RL, this approach achieves superior naturalness and stability compared to adversarial priors. Despite their success, these generative prior-based approaches still require the policy to solve a convoluted, multi-objective problem of imitation, stabilization, and task fulfillment. Our work seeks to simplify this problem through a novel decoupled approach.
\section{methodology}

\begin{figure*}[t]
    \vspace{2mm}
    \centering
    \includegraphics[width=\textwidth]{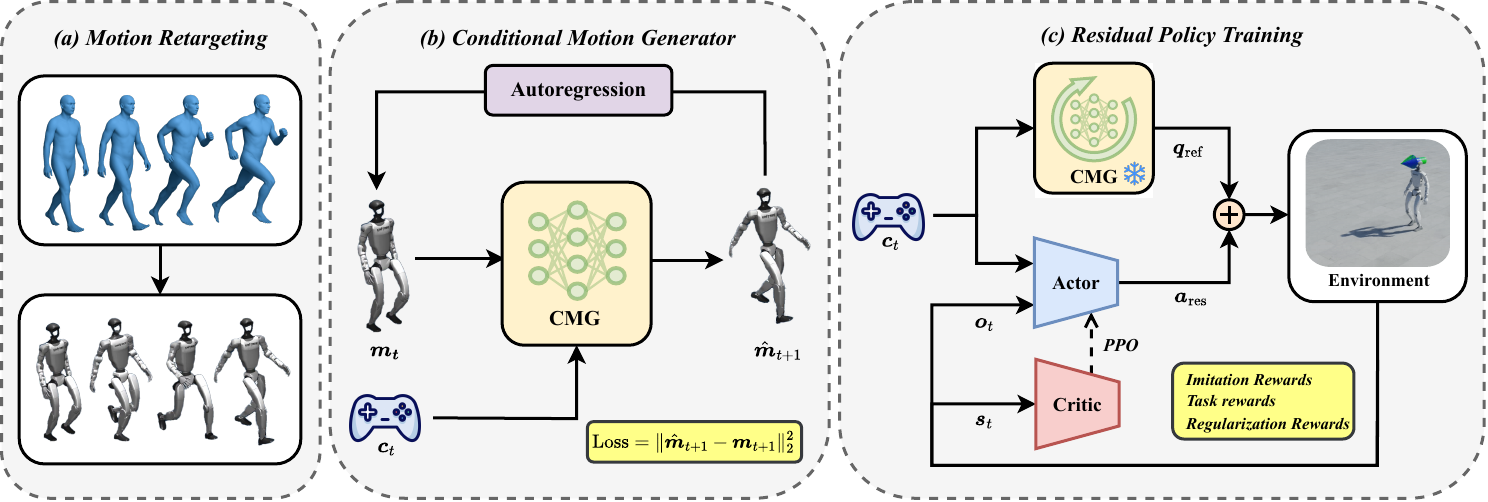}
    \caption{Overview of the RuN framework. (a) Motion Retargeting: Raw human motions are converted into a kinematically feasible reference dataset through optimization, filtering, and augmentation. (b) Conditional Motion Generator: An autoregressive CMG is trained on the curated dataset to serve as a motion prior, generating natural kinematic trajectories from user commands. (c) Residual Policy Training: A residual policy learns dynamic corrections via reinforcement learning, which are added to the CMG's output to ensure stable and task-oriented execution in a physics simulator.}
    \label{pipeline}
\end{figure*}

To enable humanoid robots to generate natural and dynamic motions, we propose RuN, a novel decoupled residual learning framework whose architecture is illustrated in Fig.~\ref{pipeline}. This framework decomposes the end-to-end motion control problem into two core modules: an autoregressive CMG and a residual policy.

Specifically, we first leverage a large-scale human motion dataset to train an autoregressive CMG offline. This generator is designed to capture the intrinsic patterns of human movement, serving as a powerful motion prior. Subsequently, during online interaction with the simulation environment, we freeze the parameters of the motion prior and train an auxiliary, lightweight residual policy network. The output of this network, a residual action, is then combined with the output from the motion prior to produce the final command for the robot.

This decoupled design ensures that the prior model preserves motion naturalness, while the residual policy focuses on adapting to complex physical dynamics and achieving precise task completion. This synergy between the expressive prior and the adaptive residual policy allows the robot to acquire complex motor skills.

\subsection{Motion Retargeting}

High-quality training data is fundamental to learning an effective motion prior. We utilize the large-scale human motion dataset, AMASS~\cite{amass}, which comprises approximately 40 hours of motion capture data parameterized by the Skinned Multi-Person Linear (SMPL) model. The SMPL model represents the human body using shape parameters $\boldsymbol\beta \in \mathbb{R}^{10}$, pose parameters $\boldsymbol\theta \in \mathbb{R}^{24 \times 3}$, and a global translation $\boldsymbol{p} \in \mathbb{R}^{3}$. These parameters can be mapped to a 3D human mesh with 6,890 vertices via the SMPL function $\mathcal{S}(\boldsymbol\beta, \boldsymbol\theta, \boldsymbol{p})$.

Due to the inherent kinematic discrepancies between the robot and a human, the raw motion data cannot be directly applied. Therefore, we employ motion retargeting to transfer the human motions into feasible reference trajectories for our robot. Following the methodology of HOVER~\cite{he2024hover}, we adopt an optimization-based approach to adjust the SMPL parameters to conform to the robot's kinematic constraints.

After retargeting, we further curate the dataset through the following steps:

\begin{itemize}
    \item \textbf{Data Filtering:} We apply heuristics to the dataset to select walking and running sequences that are physically plausible and kinematically feasible for our robot's morphology.
    \item \textbf{Data Augmentation:} To mitigate potential data biases and increase diversity, we perform horizontal mirroring on the selected sequences.
\end{itemize}

\begin{figure}[t]
    \centering
    \includegraphics[width=\linewidth]{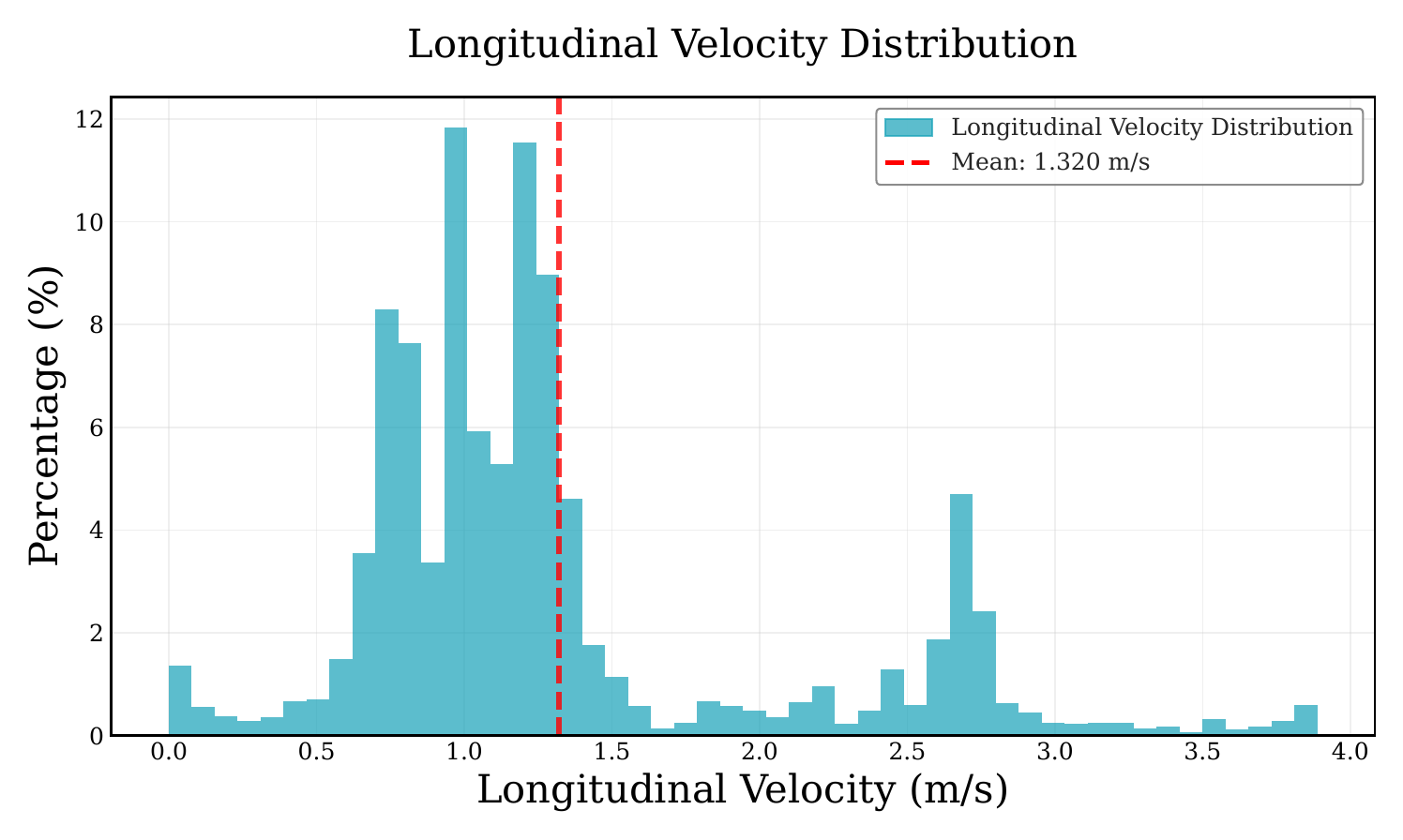}
    \caption{Longitudinal velocity distribution of our curated motion dataset after filtering and augmentation.}
    \label{fig:velocity_dist}
\end{figure}

Our pipeline yields a curated reference dataset of 400 high-quality motion sequences, which serves as the foundation for training the motion prior generator. As shown in Fig.~\ref{fig:velocity_dist}, the dataset spans a broad range of velocities, though most samples fall within 0.5~m/s to 1.5~m/s. This concentration leads to a distributional imbalance, an issue we explicitly address during the training of the CMG.

\subsection{Conditional Motion Generator}

Our motion prior is an autoregressive Conditional Motion Generator, whose core objective is to efficiently learn and generate data-driven, naturalistic motions online. Specifically, the CMG predicts the next-frame motion feature ${\hat{\boldsymbol{m}}}_{t+1}$ based on a history of $K$ motion frames, $\boldsymbol{m}_{t-K+1:t}$, and an external conditional command $\boldsymbol{c}_t$, such as a target linear velocity. The motion feature $\boldsymbol{m}_t$ is defined as the concatenation of joint positions and velocities. This process is formalized as $\hat{\boldsymbol{m}}_{t+1} = f_\theta(\boldsymbol{m}_{t-K+1:t}, \boldsymbol{c}_t)$, where $f_\theta$ is a parameterized neural network. The model is trained to minimize the Mean Squared Error (MSE) loss between the prediction and the ground-truth reference motion ${m}_{t+1}$:
\begin{equation}
    \mathcal{L}_{\text{MSE}} = \| \hat{\boldsymbol{m}}_{t+1} - \boldsymbol{m}_{t+1} \|_2^2
\end{equation}

The architecture of our CMG is inspired by the efficient decoder design in Motion VAE (MVAE)~\cite{mvae}. To enable a single model to generate diverse behaviors like walking and running, we similarly employ a Mixture-of-Experts (MoE) architecture. This architecture consists of a gating network and a 3-layer MLP backbone. The gating network computes mixing coefficients from the input, which dynamically generate the weights and biases of each backbone layer by taking a weighted sum of parameters from corresponding expert layers. Critically, departing from the MVAE framework, our CMG model completely discards the stochastic latent variable and the entire encoder structure. This design choice not only drastically reduces the parameter count but also fundamentally avoids common pitfalls of VAE-based architectures, namely training instability and posterior collapse~\cite{bowman2016generating}.

We integrate a suite of key techniques to ensure efficient and robust training. To address the distribution mismatch between training and inference, we adopt a scheduled sampling strategy~\cite{bengio2015scheduled}. The training regime gradually transitions from a teacher mode, which relies entirely on ground-truth data, to a student mode that conditions on the model's own predictions. This enhances the model's resilience to compounding errors during long-term autoregressive rollouts. Furthermore, to tackle data sparsity in scenarios like high-speed locomotion, we implement weighted data sampling by computing a distribution histogram of target velocities to up-sample less frequent examples. For data preprocessing, we apply z-score normalization to motion features, use min-max normalization for conditional commands, and inject slight Gaussian noise to improve robustness.

\begin{figure}[t]
    \vspace{2mm}
    \centering
    \includegraphics[width=\linewidth]{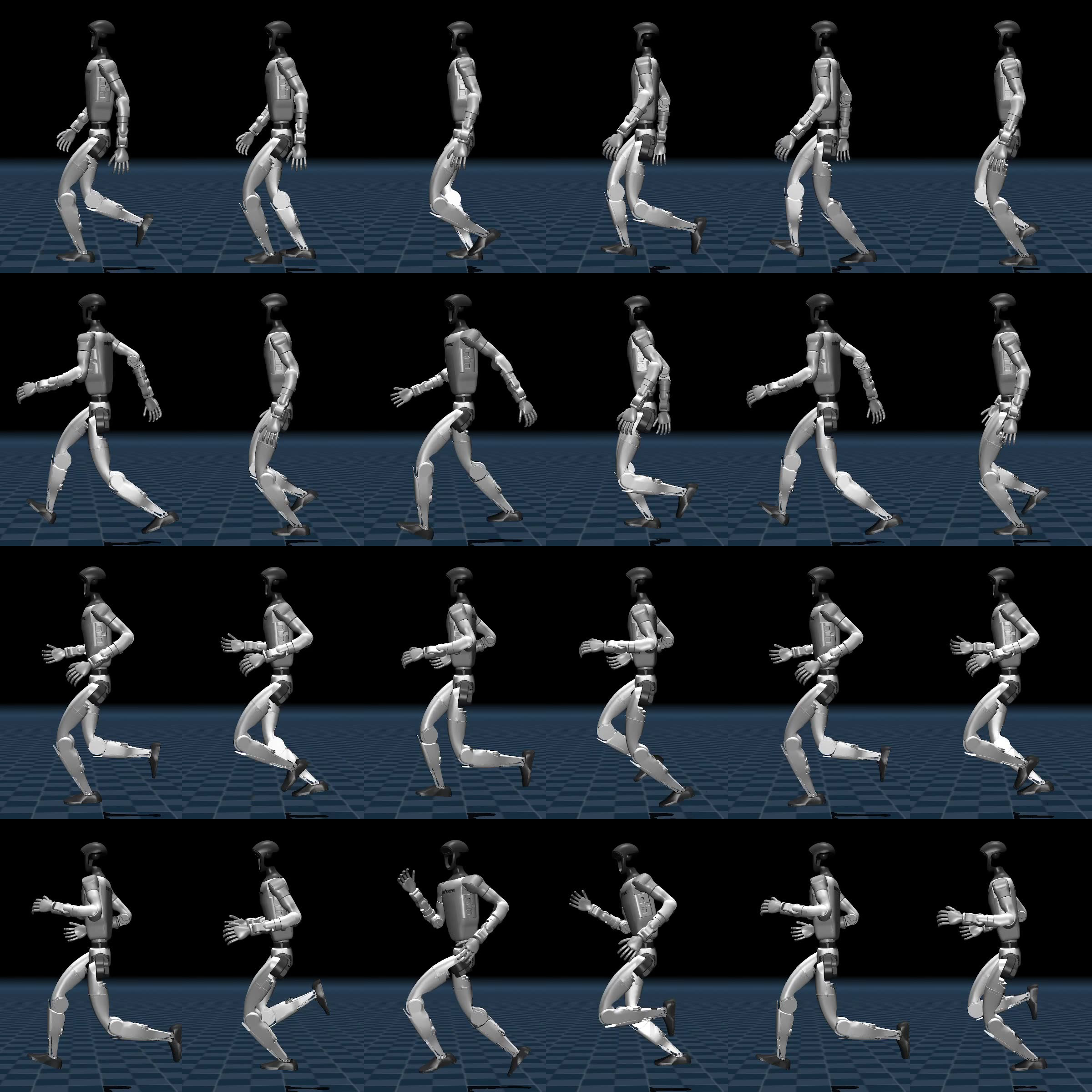}
    \caption{Visualization of motion sequences generated by our trained Conditional Motion Generator. The model produces kinematically plausible gaits that transition from walking to running in response to varying forward velocity commands (1.0~m/s, 1.5~m/s, 2.0~m/s, and 2.5~m/s).}
    \label{fig:cmg_output}
\end{figure}

In practice, we set the history length to $K=1$, simplifying the input to $(\boldsymbol{m}_t, \boldsymbol{c}_t)$. This choice is predicated on the observation that the motion feature vector, containing both joint positions and velocities, already encapsulates sufficient dynamical information, making a memoryless, first-order autoregressive model adequate for high-quality motion prediction. Once trained, the resulting model performs autoregressive prediction with minimal latency, allowing a user to smoothly control the humanoid's motion generation in real-time via velocity commands. As visualized in Fig.~\ref{fig:cmg_output}, the CMG can generate a spectrum of distinct and natural gaits corresponding to different target speeds, confirming its effectiveness as a powerful motion prior.

\subsection{Residual Policy for Locomotion}

While the CMG generates kinematically plausible motions, it operates in an open-loop fashion, oblivious to physical dynamics such as contacts, friction, or external perturbations. This lack of closed-loop feedback compromises the robot's stability, preventing it from correcting for perturbations. To address this, we introduce a residual policy trained with reinforcement learning to provide real-time, dynamic corrections.

We formalize this control task as a Partially Observable Markov Decision Process (POMDP), represented by the 6-tuple $(\mathcal{S}, \mathcal{A}, T, R, O, \gamma)$, where $\mathcal{S}$ is the set of states, $\mathcal{A}$ is the set of actions, $T$ represents the conditional transition probabilities between states, $R$ is the reward function, $O$ represents the conditional observation probabilities, and $\gamma \in [0, 1)$ is the discount factor. The policy is trained using the Proximal Policy Optimization (PPO) algorithm~\cite{schulman2017proximal} to maximize the expected future discounted reward:
\begin{equation}
    J(\pi_\theta)=\mathbb{E}_{\tau\sim\pi_\theta}\left[\sum_{t=0}^{\infty}\gamma^t r_t\right]
\end{equation}

\noindent\textbf{Decoupled Control via Residual Actions.} In contrast to end-to-end approaches, we leverage the CMG as a strong motion prior to decouple the control problem. This framework decouples the motion's kinematic style, dictated by the CMG's reference pose $\boldsymbol q_{\text{ref},t}$, from its underlying dynamics and stability, which are managed by a learned residual policy $\pi_\theta(\boldsymbol a_{\text{res},t}|\boldsymbol o_t)$. This separation significantly narrows the exploration space. The final position target for the robot's low-level PD controller, $q_{\text{target},t}$, is constructed by summing the reference pose with the corrective action from the policy:
\begin{equation}
    \boldsymbol q_{\text{target},t} =\boldsymbol q_{\text{ref},t} + \boldsymbol a_{\text{res},t}
\end{equation}

\noindent\textbf{Asymmetric State and Observation Spaces.} We employ an asymmetric actor-critic architecture~\cite{pinto2018asymmetric}. The observation $\boldsymbol o_t \in \mathcal{O}$, which is the input to the actor (policy), is designed to contain only proprioceptive and task-related information:
\begin{equation}
    \boldsymbol o_t \triangleq [\boldsymbol \omega_t, \boldsymbol{g}_t, \boldsymbol{c}_t, \boldsymbol q_t, \dot{\boldsymbol q}_t, \boldsymbol a_{t-1}],
\end{equation}
including the base angular velocity $\boldsymbol\omega_t$, the gravity vector $\boldsymbol{g}_t$, the task command $\boldsymbol{c}_t$, joint positions $\boldsymbol q_t$, joint velocities $\dot{\boldsymbol q}_t$, and the previous action $\boldsymbol a_{t-1}$.

The full state $\boldsymbol s_t \in \mathcal{S}$ is accessible only to the critic (value function) and contains privileged information about the true system state and the CMG's intention:
\begin{equation}
    \boldsymbol s_t \triangleq [\boldsymbol o_t, \boldsymbol{v}_t , \boldsymbol{m}_{t}],
\end{equation}
where $\boldsymbol{v}_t$ is the base linear velocity and $\boldsymbol{m}_{t} \triangleq [\boldsymbol q_{\text{ref},t}, \dot{\boldsymbol q}_{\text{ref},t}]$ is the motion feature from the CMG. This information asymmetry forces the policy to learn to track and stabilize the latent motion prescribed by the CMG using only onboard sensory feedback, promoting a more robust control strategy.

\noindent\textbf{Reward Design.} Our reward function, $R(\boldsymbol s_t, \boldsymbol a_t)$, is meticulously designed as a weighted sum of multiple objectives to shape a complex behavior that balances adherence to the kinematic motion prior, task fulfillment, and physical stability. The total reward value at timestep $t$, $r_t$, is composed of three core components: a motion imitation reward $r^{\text{imitation}}_t$, a task reward $r^{\text{task}}_t$, and regularization penalties $r^{\text{reg}}_t$.
\begin{equation}
    r_t = r^{\text{imitation}}_t + r^{\text{task}}_t + r^{\text{reg}}_t
\end{equation}

\textit{Imitation.} This reward forms the crucial link between the CMG prior and the RL policy, ensuring the naturalness of the final motion. It incentivizes the robot to closely track the reference motion from the prior by rewarding both pose and velocity similarity. Each term is formulated using an exponential kernel:
\begin{align}
    r^{\text{qpos}}_t &= 1.0 \cdot \exp\left(-0.6 \| \boldsymbol q_t - \boldsymbol q_{\text{ref},t} \|^2_2\right) \\
    r^{\text{qvel}}_t &= 0.2 \cdot \exp\left(-0.5 \| \dot{\boldsymbol q}_t - \dot{\boldsymbol q}_{\text{ref},t} \|^2_2\right)
\end{align}
The final imitation reward is a weighted sum of these components, with a higher weight on position tracking.

\textit{Task.} This component drives the robot to execute the user's commands, which primarily involve tracking target linear and angular velocities. The corresponding reward terms are:
\begin{align}
    r^{\text{lin}}_t &= 2.0 \cdot \exp\left(-2\| \boldsymbol v_{xy,t} - \boldsymbol c_{xy,t} \|^2_2\right) \\
    r^{\text{ang}}_t &= 0.5 \cdot \exp\left(-\| \omega_{z,t}-c_{z,t} \|^2_2\right)
\end{align}
where $v_{xy,t}$ is the robot's current planar velocity, $\omega_{z,t}$ is the robot's current yaw rate.

\textit{Regularization.} This term is a collection of penalties designed to encourage physically plausible, stable, and smooth motions, while discouraging undesirable artifacts like high torques or foot slippage. The specific formulations, motivations, and weights for each penalty are summarized in Table~\ref{Regularization Rewards}.

\begin{table}[ht]
\centering
\renewcommand{\arraystretch}{1.2}
\resizebox{\linewidth}{!}{
\begin{tabular}{lll}
\toprule
Term & Expression & Weight \\
\midrule
Alive & $\lnot~$\text{terminated} & 1.0 \\
Termination & $\mathds{1}(\text{terminated})$ & -200.0 \\
Orientation & $\|\boldsymbol{g}_{xy}\|_2^2$ & -2.0 \\
R-P Angular Velocity & $\|\boldsymbol\omega_{xy}\|_2^2$ & -0.05 \\
Energy & $\sum_{i} |\tau_i \dot{q}_i|$ & $-1 \times 10^{-5}$ \\
Action Rate & $\|\boldsymbol{a}_t - \boldsymbol{a}_{t-1}\|_2^2$ & -0.04\\
Action Smoothness & $\|\boldsymbol{a}_t - 2\boldsymbol{a}_{t-1} + \boldsymbol{a}_{t-2}\|_2^2$ & -0.06 \\
Joint Acceleration & $\|\ddot{\boldsymbol{q}}\|_2^2$ & $-5 \times 10^{-8}$ \\
Joint Limits & $\sum_i \mathds{1}(q_i \notin [q_{i, \text{min}}, q_{i, \text{max}}])$ & -2.0 \\
Joint Deviation & $\sum_{i \in \mathcal{J}_{\text{select}}} |q_i - q_{i, \text{default}}|$ & -0.2 \\
Feet Slide & $\sum_{i \in \mathcal{B}_{\text{feet}}} \|\boldsymbol{v}_{i}\| \cdot \mathds{1}(\text{contact})$ & -1.0 \\
Feet Force & $\sum_{i \in \mathcal{B}_{\text{feet}}} \max(0, \|{f}_i\| - \tau)$ & -0.003 \\
\bottomrule
\end{tabular}
}
\caption{Regularization Rewards}
\label{Regularization Rewards}
\end{table}
\section{Experiments}

We conduct extensive experiments on the Unitree G1 humanoid robot to systematically evaluate RuN. The experiments are designed to address three core research questions:

\begin{itemize}
    \item \textbf{Q1 (Effectiveness \& Efficiency):} Does our framework achieve state-of-the-art performance in motion naturalness and task completion, while also demonstrating superior training efficiency compared with leading learning-based methods?
    \item \textbf{Q2 (Component Analysis):} How do our key architectural choices—specifically the residual learning mechanism and the Asymmetric Actor-Critic (AAC) paradigm—contribute to the framework's overall success?
    \item \textbf{Q3 (Generalization):} Can a policy trained efficiently in simulation be directly deployed on the physical robot and exhibit natural, agile locomotion?
\end{itemize}

\subsection{Experimental Setup}
\label{sec:setup}

\noindent\textbf{Training Details.} All policies are trained in the NVIDIA Isaac Lab simulation platform. The control loop runs at 50~Hz with a physics step of 200~Hz. We use Proximal Policy Optimization (PPO)~\cite{schulman2017proximal} to train the residual policy network. The policy follows a user-specified command vector $\boldsymbol{c}_t = [v_x, v_y, \omega_z]$, covering a wide range of behaviors: longitudinal velocity $v_x \in [0, 2.5]$~m/s, lateral velocity $v_y \in [-0.3, 0.3]$~m/s, and yaw rate $\omega_z \in [-0.5, 0.5]~\text{rad/s}$. The policy and value networks each use a two-layer LSTM with 256 hidden units per layer. Training converges in about 8 hours on a single NVIDIA RTX 4090 GPU.

To reduce the sim-to-real gap, we train with extensive domain randomization, varying a wide range of dynamics parameters so that the learned policy remains effective despite the discrepancies between simulation and reality. The parameters and ranges are summarized in Table~\ref{tab:dr_params}.

\noindent\textbf{Metrics.} We evaluate motion quality and task performance using the following metrics:
\begin{itemize}
    \item \textbf{Fréchet Inception Distance (FID):} Measures motion naturalness as the Fréchet distance between the feature distributions of generated motions and a reference human-motion dataset~\cite{heusel2017gans,zhang2023generating}. Lower is better.
    \item \textbf{Reconstruction Loss ($\mathcal{L}_{\text{rec}}$):}Per-joint mean-squared error between the generated motion and its paired dataset ground-truth motion under matched commands, averaged over a grid of commanded velocities. Lower is better.
    \item \textbf{Imitation Errors ($E_{{qpos}}$, $E_{{qvel}}$):} Per-joint mean errors in joint positions ($E_{{qpos}}$) and joint velocities ($E_{{qvel}}$) between generated motion and motion prior under the same command. Lower is better.
    \item \textbf{Velocity Tracking Error ($E_{{vel}}$):} Mean absolute error between robot’s linear velocity and the user-commanded velocity over time. Lower is better.
\end{itemize}

\begin{table}[t]
\vspace{2mm}
\centering
\renewcommand{\arraystretch}{1.2}
\resizebox{0.8\linewidth}{!}{
\begin{tabular}{ll}
\toprule
{Parameter} & {Value} \\
\midrule
Static Friction & $\mathcal{U}(0.6, 1.0)$ \\
Dynamic Friction & $\mathcal{U}(0.4, 0.8)$ \\
Robot Mass & $\mathcal{U}(25.0, 35.0) \text{ kg}$ \\
PD Controller P Gain & $\mathcal{U}(0.8, 1.2) \times \text{default}$ \\
PD Controller D Gain & $\mathcal{U}(0.8, 1.2) \times \text{default}$ \\
Motor Friction & $\mathcal{U}(0.7, 1.3) \times \text{default}$ \\
Control Latency & $\mathcal{U}(0, 20) \text{ ms}$ \\
\bottomrule
\end{tabular}
}
\caption{Domain Randomization Parameters}
\label{tab:dr_params}
\end{table}

\subsection{Comparative Analysis (Q1)}

To validate the effectiveness of our method, we compare it against three representative baselines:
\begin{itemize}
    \item \textbf{Humanoid-Gym~\cite{gu2024humanoid}:} A standard reinforcement learning baseline trained from scratch without leveraging any motion priors.
    \item \textbf{AMP~\cite{amp}:} A classic imitation learning approach that uses an adversarial discriminator to enforce a natural motion style.
    \item \textbf{GMP~\cite{gmp}:} A state-of-the-art method that employs a pre-trained generative model as a motion prior and guides the policy via a tracking reward.
\end{itemize}

\noindent\textbf{Motion Generator Comparison.} We first isolate the motion prior and evaluate it as a standalone generator (Table~\ref{tab:gen_only_compare}). Compared to {GMP}, our {CMG} achieves lower FID and reconstruction loss $\mathcal{L}_{\text{rec}}$, indicating that CMG can produce more natural and accurate motions.

\noindent\textbf{Policy Comparison.} As shown in Table~\ref{tab:comparison}, our method achieves significantly superior performance in motion naturalness (FID) and imitation accuracy ($E_{qpos}$, $E_{qvel}$). While Humanoid-Gym exhibits a slightly lower velocity tracking error, it comes at the cost of extremely unnatural motion, as indicated by its high FID and imitation error scores. In contrast, our framework strikes an excellent balance between task performance and motion naturalness.

\begin{figure}[h]
    \centering
    \includegraphics[width=\linewidth]{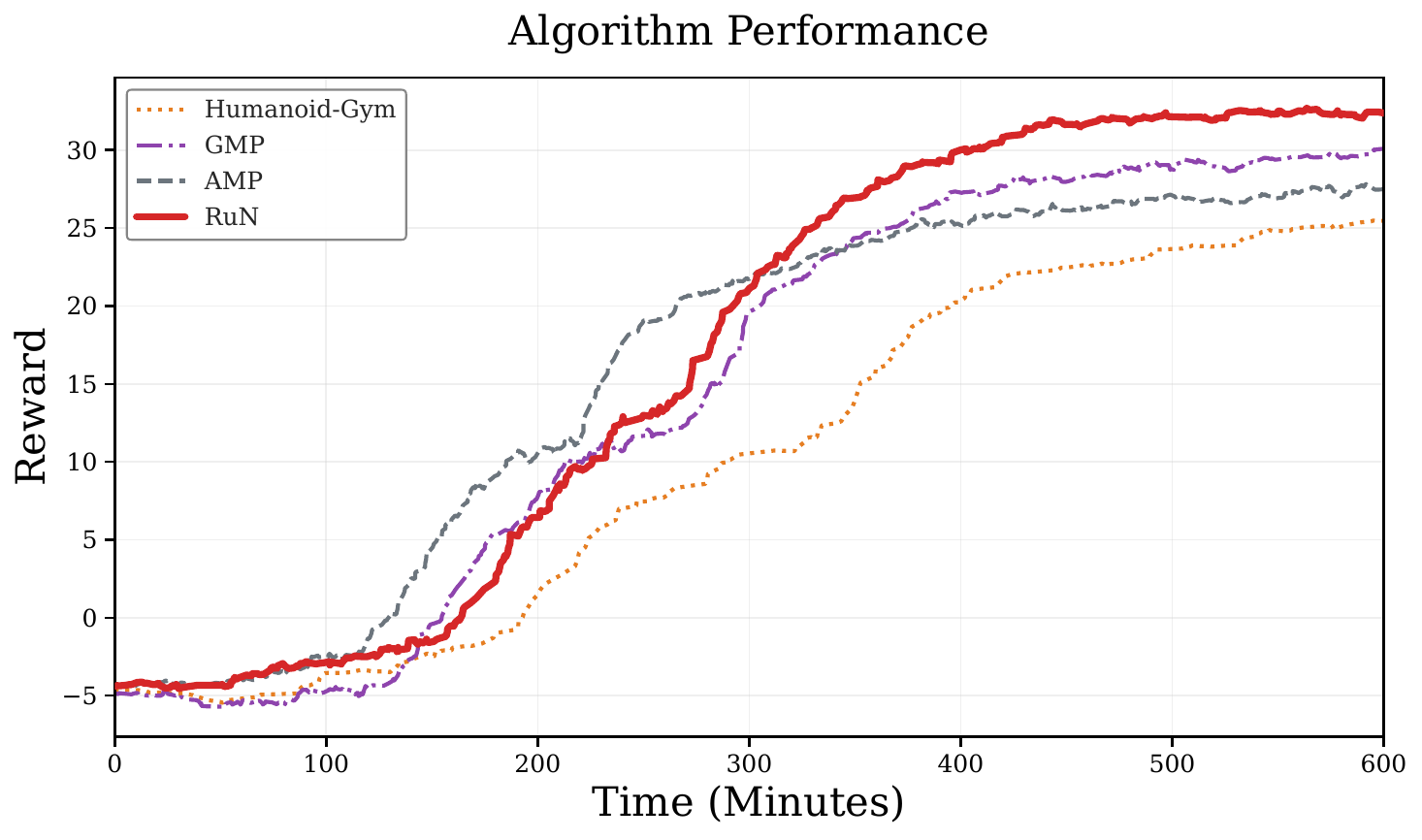}
    \caption{Performance comparison of different algorithms. This figure shows the mean episode reward over training time. Curves are smoothed for visual clarity.}
    \label{fig:learning_curves}
\end{figure}

\noindent\textbf{Training Efficiency.} In addition to final performance, training efficiency is a key indicator of a framework's practical viability. As illustrated by the learning curves in Fig.~\ref{fig:learning_curves}, our method not only achieves a higher final reward but also converges significantly faster than competing approaches. We attribute this superior efficiency to the residual learning mechanism, which constrains the exploration space to a more effective region.

\begin{table}[t]
\centering
\renewcommand{\arraystretch}{1.2}
\resizebox{0.7\linewidth}{!}{
\begin{tabular}{l|ccccc}
\toprule
Motion Generator & FID$\downarrow$ & $\mathcal{L}_{\text{rec}}\downarrow$ \\
\midrule
GMP & 0.6637 & 0.2514  \\
\textbf{CMG} & \textbf{0.5283} & \textbf{0.1556}  \\
\bottomrule
\end{tabular}}
\caption{Motion generator comparison.}
\label{tab:gen_only_compare}
\end{table}

\begin{table}[t]
\vspace{2mm}
\centering
\renewcommand{\arraystretch}{1.2}
\resizebox{\linewidth}{!}{
\begin{tabular}{l|cccc}
\toprule
{Method} & {FID $\downarrow$} & {$E_{{qpos}}$ $\downarrow$} & {$E_{{qvel}}$ $\downarrow$} & {$E_{{vel}}$ $\downarrow$} \\
\midrule
Humanoid-Gym & 3.8654 & 8.9531 & 53.2745 & \textbf{0.2732} \\
AMP & 2.8204 & 6.3817 & 46.3294 & 0.2895 \\
GMP & 1.1874 & 4.7215 & 45.9029 & 0.3172 \\
\textbf{RuN} & \textbf{0.8753} & \textbf{3.8321} & \textbf{36.7820} & 0.2873 \\
\bottomrule
\end{tabular}
}
\caption{Performance comparison with baseline methods.}
\label{tab:comparison}
\end{table}

\begin{figure*}[t]
    \vspace{2mm}
    \centering
    \includegraphics[width=\textwidth]{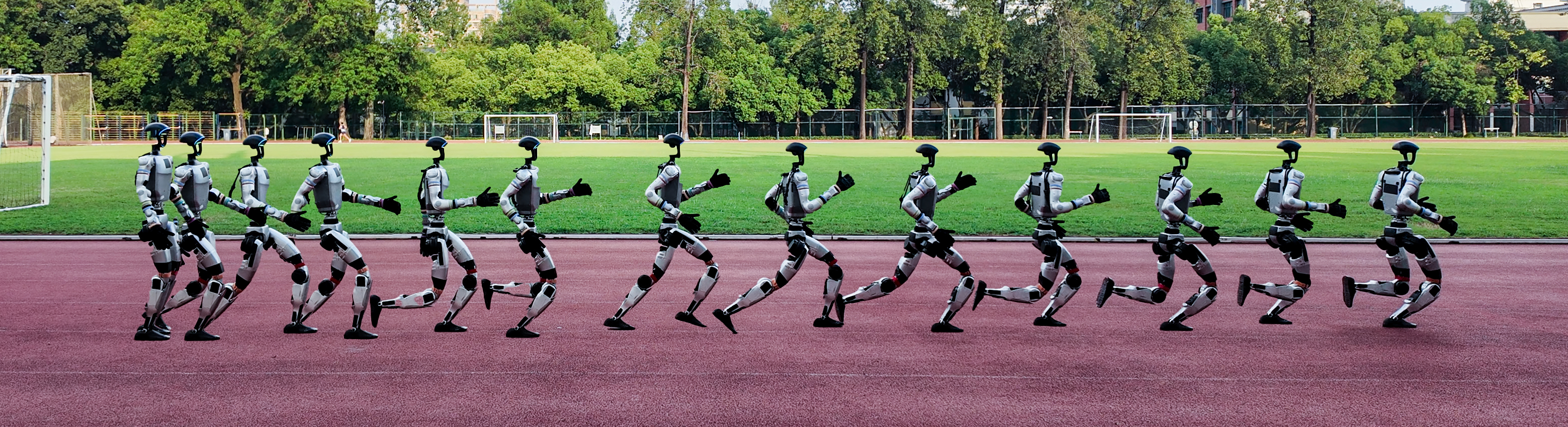}
    \caption{Snapshots of our policy deployed on the Unitree G1 robot. The robot demonstrates natural and stable locomotion, smoothly transitioning from walking to running in the real world.}
    \label{walk2run_real}
\end{figure*}

\subsection{Ablation Studies (Q2)}

To dissect RuN and assess the contribution of each component, we conduct a series of ablation studies. We remove key mechanisms and reward terms, with the results summarized in Table~\ref{tab:ablation}.

\begin{itemize}
    \item \textbf{RuN w/o CMG:} To demonstrate the crucial role of our proposed motion generator, we remove the CMG module. The policy is trained with the full reward function but without any reference motion as a prior.
    \item \textbf{RuN w/o Residual:} We replace our residual learning mechanism with a tracking policy. The policy learns the full action and is guided by a tracking reward, a paradigm conceptually similar to GMP.
    \item \textbf{RuN w/o AAC:} We remove the Asymmetric Actor-Critic mechanism, providing the critic with the same limited observations as the actor.
    \item \textbf{Reward Ablations:} We individually remove the key reward components: the joint position reward ($r^{qpos}$), the joint velocity reward ($r^{qvel}$), and the task reward ($r^{task}$).
\end{itemize}

The ablation results in Table~\ref{tab:ablation} not only confirm the necessity of each component but also reveal their distinct and synergistic roles. The analysis highlights a clear decoupling between motion quality and task fulfillment. For instance, removing the CMG leads to better velocity tracking at the cost of a drastic drop in motion naturalness, as the policy over-optimizes for the task. Conversely, taking away the task reward clearly improves motion imitation ability but causes the task to fail. The joint position reward is shown to be the most critical factor for maintaining a coherent human-like posture, with its removal causing a catastrophic failure in motion quality. Finally, the residual learning and Asymmetric Actor-Critic mechanisms are essential for our method, enabling the policy to achieve SOTA performance in both naturalness and task execution simultaneously.

\begin{table}[t]
\centering
\renewcommand{\arraystretch}{1.2}
\resizebox{\linewidth}{!}{
\begin{tabular}{l|cccc}
\toprule
{Method} & {FID $\downarrow$} & {$E_{{qpos}}$ $\downarrow$} & {$E_{{qvel}}$ $\downarrow$} & {$E_{{vel}}$ $\downarrow$} \\
\midrule
RuN w/o CMG & 2.6571 & 5.7661 & 48.2670 & \textbf{0.2759} \\
RuN w/o Residual & 1.0845 & 4.7356 & 43.2875 & 0.3068 \\
RuN w/o AAC & 1.2347 & 4.9384 & 44.3326 & 0.3012 \\
RuN w/o $r^{{qpos}}$ & 4.0586 & 8.7133 & 50.5399 & 0.2915 \\
RuN w/o $r^{{qvel}}$ & 1.0274 & 4.5692 & 46.7213 & 0.2964 \\
RuN w/o $r^{{task}}$ & 0.9542 & 3.8870 & 37.0167 & 1.6871 \\
\textbf{RuN} & \textbf{0.8753} & \textbf{3.8321} & \textbf{36.7820} & {0.2873} \\
\bottomrule
\end{tabular}
}
\caption{Ablation study results.}
\label{tab:ablation}
\end{table}

\subsection{Real-World Deployment (Q3)}

To finally validate the effectiveness of our framework, we directly deployed the policy trained in simulation onto the physical Unitree G1 robot. The deployment proved highly successful, with the robot exhibiting exceptional performance that was highly consistent with the simulation results. It stably executed continuous velocity commands from a standstill up to a 2.5~m/s run, maintaining dynamic balance across the entire range. During acceleration and deceleration, the robot demonstrated smooth and natural transitions between walking and running gaits, without any noticeable pauses or jerky motions. As visualized in Figure~\ref{walk2run_real}, the robot's posture, cadence, and coordination closely resemble human motion, successfully overcoming the stiffness often associated with traditional robotic gaits. These real-world experiments provide compelling evidence of our framework's effectiveness and its potential to solve the challenging problem of humanoid locomotion control.
\section{conclusion}

We introduced RuN, a decoupled residual learning framework that separates kinematic style from dynamic implementation for unified humanoid locomotion. A CMG supplies a high-quality, command-conditioned motion prior, while a lightweight residual policy provides dynamic corrections. Deployed on a Unitree G1, the controller delivers continuous locomotion from 0 to 2.5~m/s with smooth walk–run transitions, outperforming prior methods in motion naturalness and velocity tracking while training more efficiently. These results highlight RuN’s effectiveness and potential in humanoid locomotion, and we plan to extend this approach to uneven terrain and tighter perception–control coupling in future work.


\bibliographystyle{IEEEtran}
\addtolength{\textheight}{-16cm}
\bibliography{refs}

\end{document}